\documentclass{vgtc}                          

\graphicspath{{figures/}{pictures/}{images/}{./}} 

\usepackage{times}                     

\usepackage{tabu}                      
\usepackage{booktabs}                  
\usepackage{lipsum}                    
\usepackage{mwe}                       

\usepackage{graphicx}
\usepackage{subfigure} 

\usepackage{amsmath,amsfonts} 

\usepackage{multirow} 
\usepackage{makecell} 
\usepackage[table,svgnames]{xcolor} 
\usepackage{colortbl} 
\usepackage{diagbox} 
\usepackage{arydshln} 
\usepackage{tabularx}
\usepackage{booktabs} 
\usepackage[linesnumbered,ruled]{algorithm2e} 

\colorlet{colorFst}{LightGreen}       
\colorlet{colorSnd}{cyan} 
\colorlet{colorTrd}{LightGrey!65}      
\colorlet{colorLow}{darkgray!30}    

\colorlet{cellcolorFst}{LightGreen!35}       
\colorlet{cellcolorSnd}{LightBlue!75} 
\colorlet{cellcolorTrd}{LightGrey!65}      
\colorlet{colorLow}{darkgray!70}    

\newcommand{\comments}[1]{{\textcolor{colorLow}{#1}}}

\newcommand{\cellfs}{\cellcolor{cellcolorFst}\bf}   
\newcommand{\cellnd}{\cellcolor{cellcolorTrd}}      

\newcommand\ours{SplatLoc}
\newcommand\myvspace{\vspace{0mm}}

\usepackage{mathptmx}                  
\DeclareMathAlphabet{\mathcal}{OMS}{cmsy}{m}{n}
\DeclareSymbolFont{largesymbols}{OMX}{cmex}{m}{n}
\let\sum\relax
\DeclareSymbolFont{CMlargesymbols}{OMX}{cmex}{m}{n}
\DeclareMathSymbol{\sum}{\mathop}{CMlargesymbols}{"50}

\onlineid{1146}

\vgtccategory{Research}

\vgtcinsertpkg

\title{\ours: 3D Gaussian Splatting-based Visual Localization for Augmented Reality}

\author{Hongjia Zhai$^1$, Xiyu Zhang$^1$, Boming Zhao$^1$, Hai Li$^2$, Yijia He$^2$, Zhaopeng Cui$^1$, Hujun Bao$^1$, and Guofeng Zhang$^1$\thanks{Corresponding author: Guofeng Zhang.}
}
\affiliation{$^1$State Key Lab of CAD\&CG, Zhejiang University \quad $^2$RayNeo}

\teaser{
  \centering
  \includegraphics[width=\linewidth]{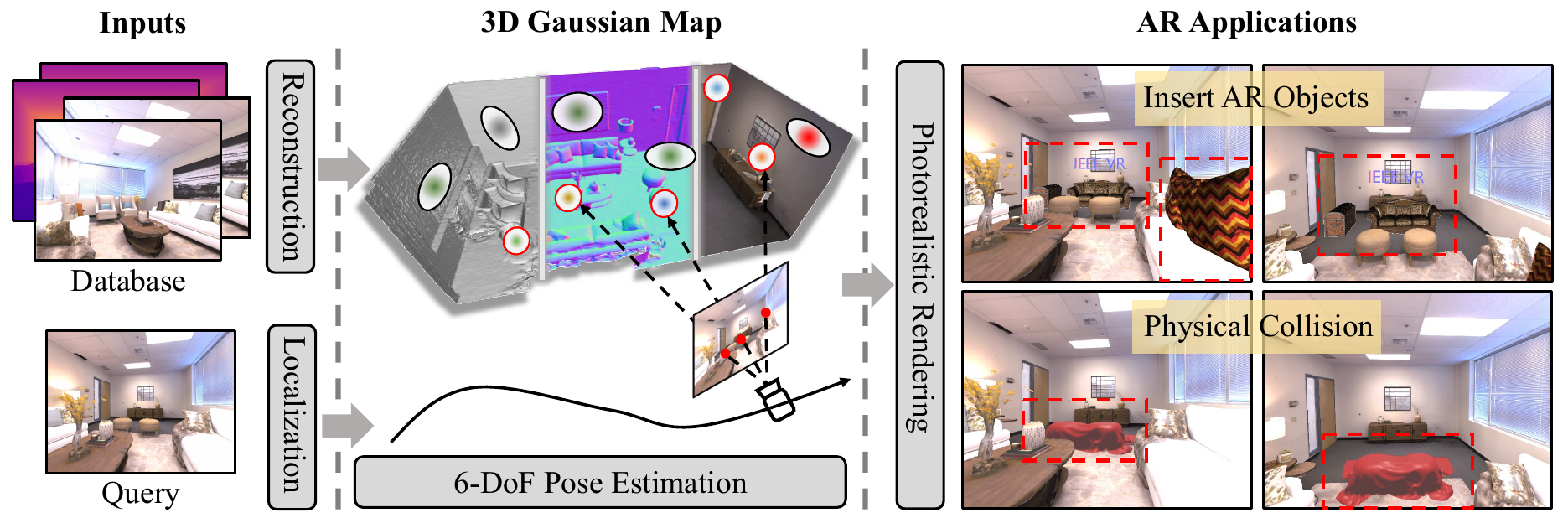}
  \caption{We present \ours, an efficient and novel visual localization approach designed for Augmented Reality (AR). As illustrated in the figure, our system utilizes monocular RGB-D frames to reconstruct the scene using 3D Gaussian primitives. Additionally, with our learned unbiased 3D descriptor fields, we achieve 6-DoF camera pose estimation through precise 2D-3D feature matching. We demonstrate the potential AR applications of our system, such as virtual content insertion and physical collision simulation. 
  We highlight virtual objects with red boxes. For AR demos, please refer to our supplementary video.}
  \label{fig:teaser}
}

\abstract{
Visual localization plays an important role in the applications of Augmented Reality (AR), which enable AR devices to obtain their 6-DoF pose in the pre-build map in order to render virtual content in real scenes. However, most existing approaches can not perform novel view rendering and require large storage capacities for maps. To overcome these limitations, we propose an efficient visual localization method capable of high-quality rendering with fewer parameters. Specifically, our approach leverages 3D Gaussian primitives as the scene representation. To ensure precise 2D-3D correspondences for pose estimation, we develop an unbiased 3D scene-specific descriptor decoder for Gaussian primitives, distilled from a constructed feature volume. Additionally, we introduce a salient 3D landmark selection algorithm that selects a suitable primitive subset based on the saliency score for localization. We further regularize key Gaussian primitives to prevent anisotropic effects, which also improves localization performance. Extensive experiments on two widely used datasets demonstrate that our method achieves superior or comparable rendering and localization performance to state-of-the-art implicit-based visual localization approaches. Project page: \href{https://zju3dv.github.io/splatloc}{https://zju3dv.github.io/splatloc}.
} 


\begin{document}



\firstsection{Introduction}
\label{sec:intro}
\maketitle

Visual localization is a critical technique that enables mobile devices or head-mounted displays to estimate the camera's 6-Degree-of-Freedom (6-DoF) pose relative to a pre-built 3D map. 
It plays an essential part in various Augmented Reality (AR) applications. For example, the visual localization approach can provide global 6-DoF pose information of the AR devices, which can be used to render virtual content in the real environment and facilitate the interaction of users with the physical space.

Generally, classical visual localization methods can be classified into two categories: regression-based and feature-based approaches.
Regression-based approaches~\cite{mapnet,src,posenet,scrnet,scrnet-id} usually use convolutional neural networks (CNN) to extract the high-level contextual feature of the image and encode the geometry information (\textit{e.g.}, absolute pose, and scene coordinate) of the reconstructed environment.
PoseNet~\cite{posenet} and SCRNet~\cite{scrnet} are the representative works of directly regressing the pose or 3D coordinates of the pixels from the extracted feature of a single image.
However, due to the lack of geometric constraints, these methods often lag behind feature-based approaches in terms of accuracy.
Feature-based approaches~\cite{hloc,bgnet,pnerfloc} usually build a structure-based scene map beforehand (\textit{e.g.}, 3D point cloud models) and associate each map primitive with one or more 3D descriptors.
Those 3D consistent descriptors are usually obtained by performing multi-view fusion on the hand-crafted feature~\cite{sift} or learning-based keypoint descriptors~\cite{superpoint,r2d2} detected from 2D images.
The detected 2D points in the query image can be matched against 3D descriptors to obtain 2D-3D correspondences for robust pose estimation~\cite{p3p,epnp}.
The localization performance of feature-based approaches is also decided by the repeatable and discriminative power of extracted descriptors.
However, limited by the way of scene representation, these classical localization approaches can not perform photorealistic rendering, which is an essential part of AR applications.

In recent years, Neural Radiance Fileds (NeRF)~\cite{mildenhall:2020:nerf} and 3D Gaussian Splatting (3DGS)~\cite{kerbl3Dgaussians} have emerged as the new paradigm for neural implicit scene representation.
These paradigms use implicit representation (\textit{e.g.}, multilayer perception~\cite{sucar:2021:imap}, parametric encodings~\cite{Vox-Surf,mueller2022instantngp,nis-slam}) or explicit primitives (\textit{e.g.}, points~\cite{xu2022pointnerf}, 2D/3D Gaussians~\cite{Huang2DGS2024,kerbl3Dgaussians}) to represent the scene property and achieve satisfactory performance of high-quality rendering and geometry reconstruction.
Benefiting from differentiable NeRF-style volumetric rendering~\cite{volume_rendering} and point-based alpha-blending~\cite{max1995optical}, neural-based approaches can perform parameter optimization in an end-to-end way without 3D supervision.
Inspired by~\cite{mildenhall:2020:nerf,xu2022pointnerf}, some works~\cite{uncertainty-loc,inerf,ma2024continuous,vox-fusion,vox-fusion++,pnerfloc} use neural implicit representation to reconstruct the scene and perform pose estimation.
iNeRF~\cite{inerf} is the first work that refines the 6-DoF pose via photometric error minimization between the query image and rendering results of the pre-trained NeRF model.
NeRF-SCR~\cite{uncertainty-loc} and LENS~\cite{lens} are the representative works that combine regression-based visual localization with neural radiance field.
They train a scene-specific NeRF model to synthesize high-quality novel views to cover the whole scene space, providing additional training data for the optimization of their scene coordinate regress network.
Similarly, the localization performance of these NeRF-aided regression approaches is also not competitive due to the lack of geometric constraints.
To impose the geometric constraints, the feature-based method, PNeRFLoc~\cite{pnerfloc}, represents the scene with explicit structure~\cite{xu2022pointnerf} and associates each point in the map with learning-based descriptor~\cite{r2d2}.
Compared with~\cite{uncertainty-loc,lens}, PNeRFLoc~\cite{pnerfloc} can achieve better localization performance and generalizability.
However, PNeRFLoc, like traditional feature-based methods, requires explicit storage of point-wise features, which results in significant memory usage, making it impractical for mobile devices with limited storage.

To overcome the aforementioned limitations, we propose an efficient and novel visual localization method that achieves better performance with fewer model parameters, suitable for both localization and high-quality novel view rendering.
Specifically, to reduce the model parameters, we do not store point-wise descriptors explicitly. Instead, we construct the feature volume from multi-view 2D feature maps and distill it to a scene-specific 3D feature decoder, which can avoid the descriptor bias of Gaussian primitives introduced by alpha-blending.
We then propose an efficient salient 3D landmark selection algorithm to reduce the computational overhead of 2D-3D matching caused by a large number of Gaussian primitives.
Finally, we perform position and scaling regularization for key Gaussian primitives to reduce the 3D center shift.
Overall, the specific contributions of our proposed approach are summarized as follows:
\begin{itemize}
    \item{We propose an efficient and novel visual localization approach based on 3D Gaussian primitives that achieves accurate localization performance and high-quality, fast rendering with fewer parameters.}
    \item{We introduce an unbiased 3D descriptor learning strategy for precise 2D keypoints and 3D Gaussian primitives matching, using a scene-specific 3D feature decoder to regress the feature volume from multi-view feature maps.}
    \item{We develop an effective salient 3D landmark selection algorithm to reduce the number of primitives used for localization. Additionally, to mitigate the Gaussian primitive center shift induced by photometric rendering loss, we apply regularization on the location and scale of key Gaussian primitives.}
    \item{We conduct extensive experiments to demonstrate the state-of-the-art and comparable performance of visual localization and high-quality novel view rendering.}
\end{itemize}

The rest of our paper is structured as follows: In~\cref{sec:related_work}, we provide a review of related works to contextualize our research.
Next, in \cref{sec:method}, we explain each key component in our reconstruction pipeline.
Subsequently, in \cref{sec:exp}, we evaluate the performance of our system through various synthetic and real-world scenes.
Finally, the conclusion and limitation are drawn in \cref{sec:conclusion}.
\section{Related Works}
\label{sec:related_work}
In this section, we review the works most relevant to the proposed method, including visual localization, 3D Gaussian Splatting, and 3D feature field.

\subsection{Visual Localization}
Visual localization is one of the most commonly used localization techniques for applications~\cite{imtooth,ming2024benchmarking} of AR/VR, which can estimate the accurate 6-DoF pose of the camera.
Generally, visual localization approaches can be divided into feature-based and regression-based methods.
Regression-based approaches represent the map in an implicit way, which encodes the geometry information of the scene in the convolutional neural networks to regress the absolute pose~\cite{mapnet,posenet} and scene coordinates~\cite{src,scrnet}.
However, both pose and coordinate regression methods need a large amount of 3D training data to make the scene-specific CNN model convergence and are not competitive in terms of localization accuracy.
Feature-based approaches~\cite{Li:2021:BDLoc,hloc,bgnet} obtain the pose via performing feature matching between the 2D keypoints of the query image and 3D points in the sparse Structure-from-Motion (SfM) model.
According to the estimated 2D-3D correspondences, the pose of the query image can be computed via the Perspective-n-Point (PnP) algorithm~\cite{epnp}.
The localization performance of feature-based methods relies on the discriminative ability of the hand-crafted~\cite{sift} or learning-based keypoint descriptors~\cite{superpoint,r2d2}.
Recently, inspired by the neural implicit representation~\cite{mildenhall:2020:nerf}, some works have used neural implicit representation to aid visual localization.
For example, \cite{uncertainty-loc} and \cite{lens} use a pre-trained NeRF model to synthesize more high-quality novel views for the training of scene regression networks.
PNeRFLoc~\cite{pnerfloc} uses the explicit point-based neural representation to employ the geometry constraints and perform 2D-3D feature matching for 6-DoF pose estimation.
However, similar to traditional feature-based methods, \cite{pnerfloc} also require a lot of memory to store the descriptors and matching graph for the sparse SfM model.

\subsection{3D Gaussian Splatting}
Recently, 3D Gaussian splatting (3DGS)~\cite{kerbl3Dgaussians} has shown great potential in 3D computer vision~\cite{chen2024pgsr,chung2024depth_gaussian,Huang2DGS2024,huang2024sc_gs,lin2024vastgaussian,ma2024shapesplat,wu20244d_gaussian,yan2024gs_slam}.
Compared to the neural radiance field (NeRF)~\cite{mildenhall:2020:nerf}, 3DGS uses an explicit representation to reconstruct the scene and enable fast and high-quality view synthesis.
Besides, 3DGS gets rid of the need for MLP and ray marching, which can accelerate the training speed for real-time rendering.
Although 3DGS can achieve high-quality novel view rendering and fast training and rendering speed, they usually lead to poor geometric accuracy, \textit{e.g}, the center of primitives are not accurately aligned with the surface.
Some works aim to control the shape of the primitives~\cite{Huang2DGS2024}, use unbiased depth rendering~\cite{chen2024pgsr}, and introduce geometric regularization during the optimization process, such as monocular depth~\cite{monogs,yan2024gs_slam}, and normals~\cite{turkulainen2024dn_splatter,xiang2024gaussianroom}.
Besides, some works~\cite{niedermayr2024compressed_3dgs,papantonakis2024reducing_memory} pay attention to reducing the memory footprint of Gaussian splatting operations.
Additionally, to enhance the scene understanding ability, \cite{qin2024langsplat,shi2024_gs_language_embed} assign an additional attribute, semantic feature, for each Gaussian primitive.
To render the high-dimensional features, they perform quantization or dimensionality reduction, which leads to performance decreases for large indoor scenes.

\begin{figure*}[ht!]
\centering
 \includegraphics[width=\linewidth]{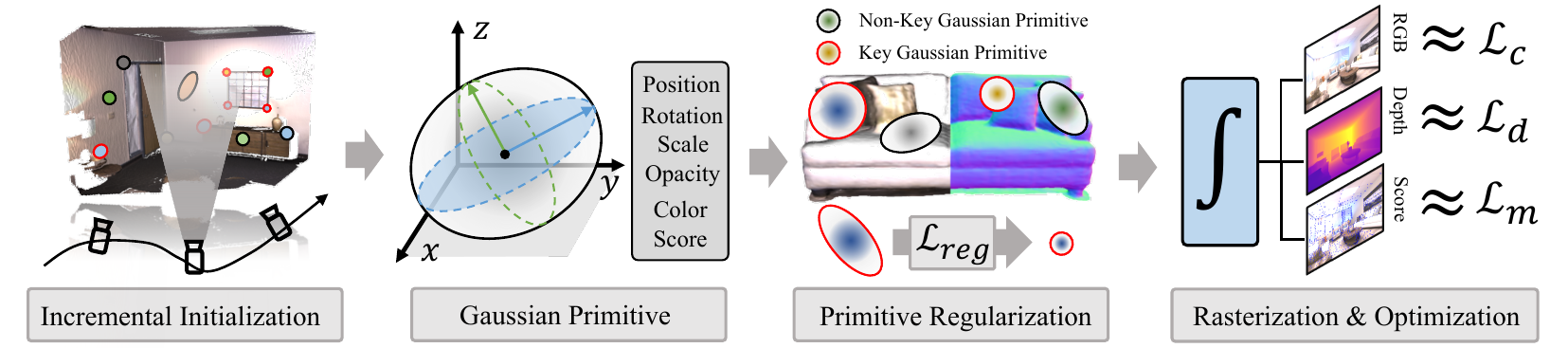} 
\caption{Reconstruction processes. We incrementally initialize the Gaussian primitives, and each primitive is associated with position $\mu$, rotation $q$, scale $s$, opacity $\sigma$, color $c$, and 3D landmark score $a$. 
For key Gaussian primitives, we perform soft isotropy and scale regularization to mitigate the anisotropic results. 
The color loss $\mathcal{L}_{c}$, depth loss $\mathcal{L}_d$, 3D landmark loss $\mathcal{L}_m$, and regularization loss $\mathcal{L}_{reg}$ are used to optimize the properties of each primitive via differentiable rasterization.}
\label{fig:reconstruction}
\end{figure*}
\subsection{3D Feature Field}
Since the development of NeRF, some works have attempted to learn consistent 3D feature fields from the 2D feature map distilled from vision or language foundation models~\cite{dino,sam,clip}.
DecomposingNeRF~\cite{decom-nerf} is the first work that adds an additional branch for the neural network to encode the semantic feature.
The high-level contextual feature learned from~\cite{Lseg,dino} can be used to distinguish the background and objects, which can be used for many editing tasks.
To reduce the memory footprint, N3F~\cite{n3f} uses Principal Component Analysis (PCA) to compress the high dimensional DINO feature into lower dimensions for efficient 3D segmentation.
Similar to NeRF-based methods, some recent works try to add feature attributes for Gaussian primitives.
However, due to the memory demands of recording high-dimensional embeddings for each primitive, it is infeasible like the way in~\cite{decom-nerf}.
To tackle those problems, Shi~\textit{et al.}~\cite{shi2024_gs_language_embed} propose an efficient quantization scheme to reduce the memory requirement of language embeddings.
Qin~\textit{et al.}~\cite{qin2024langsplat} train an auto-encoder to obtain a very low dimensional latent feature, which is associated with the Gaussian primitives.
Those latent features can be decoded into high dimensions, which can be used for localization and segmentation.
Learning feature fields via optimizing the similarity between rendered features and 2D feature maps from foundation models is not suitable for 2D-3D matching. There is a domain gap between the training and inference stages.
\section{Method}
\label{sec:method}

In this section, we introduce the reconstruction and localization pipelines of our localization approach.
In~\cref{subsec:3dgs}, we first introduce the 3D Gaussian scene representation.
Then, we introduce the unbiased 3D descriptor feature field learning (\cref{subsec:distill_feat}).
Besides, in~\cref{subsec:score_select}, we introduce the 3D landmark selection strategy for localization and model compression.
The regularization terms and overall optimization objectives of our reconstruction system are shown in~\cref{subsec:regularization} and~\cref{subsec:loss}.
Finally, the visual localization process is introduced in~\cref{subsec:localization}.

\begin{figure*}[h]
\centering
\subfigure[Blending-based biased 3D feature learning.]{
 \includegraphics[width=0.48\linewidth]{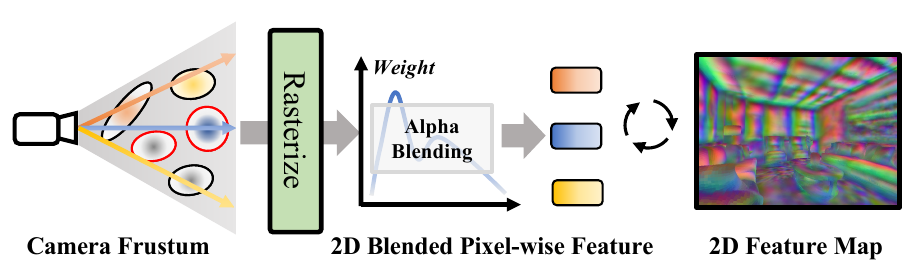} 
 \label{fig:biased_feature}
}
\subfigure[Regression-based unbiased 3D feature learning.]{
 \includegraphics[width=0.48\linewidth]{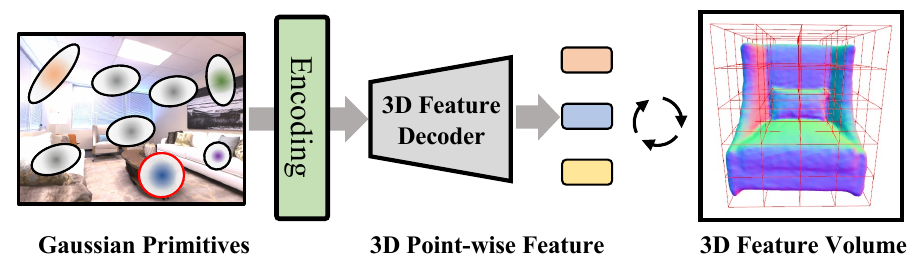} 
 \label{fig:unbiased_feature}
}
\vspace{-3mm}
\caption{Illustration of biased and unbiased 3D descriptor field learning. (a) The biased 3D feature optimization of previous works~\cite{qin2024langsplat,shi2024_gs_language_embed}, they use alpha-blending to obtain the 2D blended feature. (b) Our unbiased 3D feature learning scheme, which directly learns the 3D feature decoder from the constructed feature volume of multi-view feature maps.}
\label{fig:feat_comp}
\end{figure*}
\begin{figure*}[ht!]
\centering
 \includegraphics[width=\linewidth]{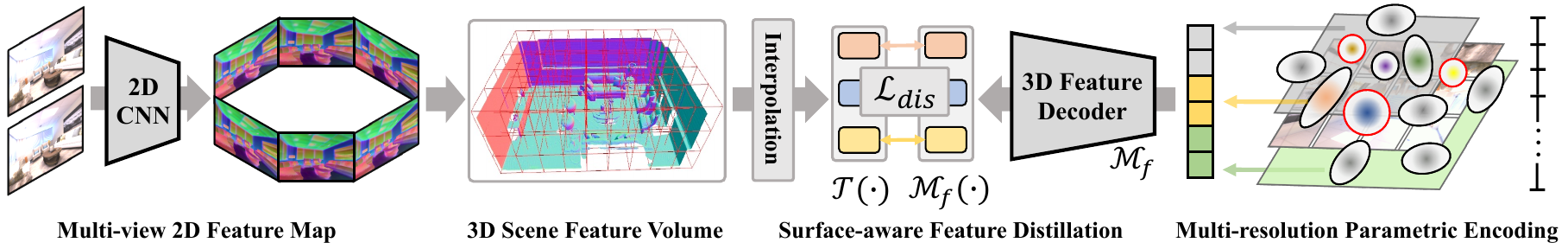} 
\caption{The pipeline of our unbiased 3D primitive descriptor learning. We first encode images based on the 2D CNN model~\cite{superpoint} to obtain the multi-view feature maps and construct the 3D scene feature volume according to the depth and pose information. To enhance the representation ability of the 3D feature decoder, we use multi-resolution parametric encoding to aid the 3D scene-specific descriptor learning. Besides, we only sample descriptors on the scene surface for effective distillation.}
\label{fig:distill_feat}
\end{figure*}
\subsection{3D Gaussian Scene Representation}
\label{subsec:3dgs}
In this part, we introduce the Gaussian primitive-based scene representation and incremental reconstruction process (\cref{fig:reconstruction}).

\myvspace\noindent\textbf{Scene Representation.}
3D Gaussian Splatting utilizes a collection of anisotropic 3D Gaussian primitives to represent the scene explicitly.
Specifically, each Gaussian primitive is parameterized by its mean $\mu\in \mathbb{R}^{3}$ and covariance matrix $\Sigma\in \mathbb{R}^{3\times3}$ defined in the world space, which is shown in the following:
\begin{equation}
G(\mu,\Sigma) = e^{-\frac{1}{2}(x-\mu)^T\Sigma(x-\mu)}.
\end{equation}
To ensure the covariance matrix $\Sigma$ maintains physical meaning during optimization, it is decomposed into a scaling matrix $S$ and a rotation matrix $R$ as proposed in~\cite{kerbl3Dgaussians}:
\begin{equation}
\Sigma = RSS^TR^T,
\end{equation}
where the scaling matrix is computed from a 3D scale vector $\textbf{s}$, $S=\text{diag}([\textbf{s}])$, and the rotation $R$ is parameterized by quaternion.

Following the approach in Zwicker~\textit{et al.}~\cite{zwicker2001ewa}, the 3D Guassians are projected into the 2D image plane for rendering.
Given the viewing transformation matrix $W$ and Jacobian $J$ of the affine approximation of the projective transformation, the covariance matrix in the camera coordinates can be computed as:
\begin{equation}
\widetilde{\Sigma} = JW \Sigma W^TJ^T.
\end{equation}
The corresponding 2D Gaussian distribution $\hat{G}(\widetilde{\mu},\widetilde{\Sigma})$ is derived from the 2D pixel location $\widetilde{\mu}$ of the 3D Gaussian primitive center and the projected covariance matrix $\widetilde{\Sigma}$.

\myvspace\noindent\textbf{Differentiable Rasterization.}
For novel view synthesis and fast rasterization-based rendering, each 3D Gaussian primitive is associated with an opacity $\sigma\in \mathbb{R}$ and a color $c\in \mathbb{R}^3$, represented using spherical harmonics (SH) coefficients.
For photo-realistic rendering, the differentiable rasterizer adopts alpha-blending~\cite{max1995optical} to render the Gaussian property into the image plane, which accumulates Gaussian property and opacity values $\sigma$ on a given pixel by traversing the ordered primitives.
Specifically, the depth and color properties can be rendered with the following equation:
\begin{equation}
    \hat{I} =\sum_{i = 1}^{N} c_i \cdot \alpha_i \cdot \prod_{j=1}^{i-1}(1-\alpha_j), \quad \hat{D} =\sum_{i = 1}^{N} d_i \cdot \alpha_i \cdot \prod_{j=1}^{i-1}(1-\alpha_j),
    \label{eq:render_rgb_depth}
\end{equation}
where $\hat{I}$, and $\hat{D}$ are the rendered color and depth, respectively. $\alpha_i=\hat{G}(\widetilde{\mu},\widetilde{\Sigma})\cdot \sigma_i$ denotes the opacity contribution of each 2D pixel, $\prod_{j=1}^{i-1}(1-\alpha_j)$ is the accumulated transmittance, and $N$ is the number of Gaussian primitives during the splatting process for a pixel.

Additionally, to identify salient 3D landmarks for accurate and efficient visual localization, each Gaussian primitive is assigned a 3D landmark probability score $a \in \mathbb{R}$, representing the likelihood of the primitive being a key landmark in 3D space. 
Also, similar to the color and depth rendering equation, we can render the 3D landmark probability score into 2D with the following equation:
\begin{equation}
    \hat{A} = \sum_{i=1}^{N} a_i \cdot \alpha_i \cdot \prod_{j=1}^{i-1}(1-\alpha_j).
    \label{eq:render_score}
\end{equation}

\myvspace\noindent\textbf{Incremental Gaussian Initialization.}
During the reconstruction process, Gaussian primitives are initialized incrementally for each keyframe.
For each incoming keyframe $\{I,D\}$, we first random sample pixels and project the sampled points into 3D space according to the camera pose, $T_{wc}$, and camera intrinsic, $K$, with the following equation:
\begin{equation}
    \mu = T_{wc} \cdot \pi^{-1} (u, D(u)),
\end{equation}
where $\pi^{-1}(\cdot)$ is the inverse of perspective projection, $u$ is the sampled pixel.
Then the Gaussian primitive is initialized with the projected 3D center $\mu$ and the color value from $I(u)$.

Additionally,using the 2D keypoint score map generated by the SuperPoint model~\cite{superpoint}, we classify Gaussian primitives with scores exceeding a defined threshold as key Gaussian primitives, while others are marked as non-key primitives. Key Gaussian primitives typically exhibit higher 3D landmark probability scores and are more reliable for localization during the inference stage.

\subsection{Unbiased 3D Descriptor Learning}
\label{subsec:distill_feat}
To achieve effective 2D-3D feature matching while compressing the scene parameters, we propose learning descriptors for each Gaussian primitive using a 3D feature decoder. 
As shown in~\cref{fig:biased_feature}, previous approaches~\cite{qin2024langsplat,shi2024_gs_language_embed} project 3D primitive feature into 2D blended feature via alpha-blending, then perform similarity optimization with CNN-generated 2D feature maps. However, such methods introduce a domain gap between the 2D CNN feature and the 3D Gaussian primitive descriptor, making them unsuitable for our task of 2D-3D feature matching.
Furthermore, due to GPU memory constraints, directly learning high-dimensional descriptors is impractical. Existing solutions that rely on dimensionality reduction~\cite{qin2024langsplat} or feature quantization~\cite{shi2024_gs_language_embed} negatively affect localization performance. To address these issues, we propose an unbiased, accuracy-preserving 3D descriptor learning method (\cref{fig:unbiased_feature}), which directly regresses the 3D feature volume without any feature quantization and dimensionality reduction.
The whole learning process is shown in~\cref{fig:distill_feat}.

\myvspace\noindent\textbf{2D Feature Map Lifting.}
To avoid the bias introduced by alpha-blending, we lift the 2D feature map into 3D feature volume and learn a 3D feature decoder to generate scene-specific descriptors for primitives.
Specifically, for each keyframe, we first use a pre-trained CNN~\cite{superpoint} to extract its 2D feature map, $f_i \in \mathbb{R}^{H\times W \times D_f}$, where $D_f$ denotes the feature map dimension. 
According to the 6-DoF camera pose and intrinsics, we then project 2D feature maps into 3D feature volume, $\mathcal{V}\in \mathbb{R}^{\{D_x \times D_y \times D_z \times D_f\}}$, where $D_x$, $D_y$, and $D_z$ are the volume dimensions based on scene size and voxel resolution.
To model geometry information, we follow~\cite{dai:2017:bundlefusion} and apply an additional TSDF volume to capture surface details from depth images.
Besides, we update the 3D feature volume by fusing multi-view observations.
Specifically, for each feature at position $(x,y,z)$ in the volume, we apply weighted average pooling for the observation from different viewpoints:
\begin{equation}
    \mathcal{V}[x,y,z] = \frac{w_i \cdot \mathcal{V}[x,y,z] + f_i(u)}{w_i + 1},
\end{equation}
where $(x,y,z)$ is the volume coordinates derived from pixel location $u$ and its corresponding depth value, and $w_i$ is the weight of the current volume when fusing the $i$-th feature map $f_i$.
So, we can obtain the feature of any spatial point in the scene through the trilinear interpolation operation.

\myvspace\noindent\textbf{Multi-resolution Parametric Encoding.}
As shown in~\cite{qin2024langsplat,sucar:2021:imap}, a single MLP or encoder-decoder module leads to a performance decrease in scene representation and feature learning.
To improve performance, we adopt multi-resolution hash feature encoding~\cite{mueller2022instantngp} to enhance the MLP decoder’s ability to represent complex scenes, as shown on the right of~\cref{fig:distill_feat}.
Specifically, for each Gaussian primitive with its 3D position $\mu$ we can obtain its multi-resolution parametric encoding $\mathcal{E}(\mu;\Theta)$ using the following equation:
\begin{equation}
    \mathcal{E}(\mu;\Theta) = [\mathcal{T}(\mu,\theta_1), \cdots, \mathcal{T}(\mu,\theta_l)],
\end{equation}
where $\Theta$ is the multi-resolution features $\Theta$ = \{$\theta_l$\}$_{l=1}^L$, and $\mathcal{T}(\cdot)$ is the trilinear interpolation operation for each resolution level.

After obtaining the multi-resolution encoding feature, $\mathcal{E}(\mu;\Theta)$, the 3D descriptor of the primitive can be obtained with the 3D feature decoder, $\mathcal{D}$, which is implemented by a shadow MLP:
\begin{equation}
     g = \mathcal{M}_{f}(\mathcal{E}(\mu;\Theta)),
     \label{eq:decode_feat}
\end{equation}
where $g\in \mathbb{R}^{D_f}$ is the decoded high-dimensional descriptor at 3D position $\mu$, which can be directly optimized with our constructed 3D feature volume $\mathcal{V}$.

\myvspace\noindent\textbf{Surface-aware Descriptor Distillation.}
Since much redundant information is stored in the feature volume $\mathcal{V}$ (\textit{e.g.}, the descriptors in empty 3D space are often invalid and inaccurate), so we only perform descriptor distillation on the scene surface manifold.
To achieve this, we first use the Marching Cubes algorithm~\cite{lorensen1998marching} to find the surface inside the feature volume.
Then, during the distillation process, we randomly sample points on the surface and use the interpolation operation to obtain their corresponding features, which are then used to optimize the multi-resolution features $\Theta$ and the 3D feature decoder $\mathcal{M}_f$.

\begin{algorithm}[ht!]
\caption{Salient 3D Landmarks Selection}
\label{alg:select_gaussian}

\LinesNumbered 
\KwIn{Primitives $\mathcal{P}$, Search radius $r$, Number $N$} 
\KwOut{Selected landmarks $\mathcal{S}$} 

\comments{// Compute saliency score for all primitives} \\
\For{$p \in \mathcal{P}$}
{
\comments{// Significance + Generalizability  + Consistency}  \\
$\mathcal{H}(p) = 2 \cdot \text{Sig}(p) + \min(2,\; \text{Gen}(p)) + \text{Geo}(p)$ \\
}
\comments{// Search $N$ suitable landmarks} \\
\While{$|\mathcal{S}| < N$}
{
\comments{// Saliency-based greedy search} \\
{$\Omega = \{p | p \in \mathcal{P}, ||x-s|| > r, \forall s \in \mathcal{S} \}$; \quad \comments{// Candidates}} \\
$p = \mathop{\arg\max} \mathcal{H}(p), \; \text{for} \; {p \in \Omega}$;  \\
\eIf{$p$ \text{exist}}
{$\mathcal{S}$ = $\mathcal{S} \cup p$;  \quad \comments{// Find the target landmark}}
{$r$ = $r / 2$; \quad \comments{// Reduce search radius}} 
}
return $\mathrm{S}$ \quad \comments{// The final selected landmarks}
\end{algorithm}

\subsection{Salient 3D Landmark Selection}
\label{subsec:score_select}
Due to the need for photo-realistic rendering, 3DGS typically produces a large number of Gaussian primitives for indoor scenes (\textit{e.g.}, around 400k points for \texttt{Room 0} in the Replica dataset).
However, many of these primitives contain redundant information that is unnecessary for visual localization. 
For mobile or lightweight devices, using all these primitives would incur significant computational costs. To address this, we propose an efficient 3D landmark selection strategy based on the primitive saliency score, reducing the number of primitives required for localization.

\myvspace\noindent\textbf{Saliency Score Computation.}
Previous works~\cite{bergamo2013leveraging,Do_2022_SceneLandmarkLoc,mera2020efficient,yang2022scenesqueezer} have reduced the number of map points based on differentiable optimization or the importance of each point. However, there is no clear standard for evaluating the distinctiveness of points for accurate pose estimation.
To reduce the number of primitives for localization, the 3D landmark selection method should find the suitable subset of Gaussian primitives from the reconstructed scene based on our defined saliency score.
To select the informative 3D landmarks for localization, we define a saliency score for each Gaussian primitive $p\in \mathcal{P}$, which is computed based on the following criteria:

\textbf{(1) Significance}:
Primitives with higher significance tend to be 3D landmarks, making them more useful for localization.
As described in~\cref{subsec:3dgs}, each Gaussian primitive $p$ is assigned a 3D landmark probability score $a$, representing the likelihood of being a landmark.
So, for the significance term, we use the learnable landmark probability score $a$ as the evaluation metric:
\begin{equation}
    \text{Sig}(p) = a.
\end{equation}

\textbf{(2) Generalizability}:
Primitives that were observed from many different viewing directions during the reconstruction are more generalizable and robust for localization. For this generalizability term, we use the largest angle between any two viewing directions as the evaluation metric which is defined in the following equation:
\begin{equation}
    \text{Gen}(p) = \max \{\arccos(\frac{(o_i - \mu)}{|| (o_i - \mu) ||} \cdot \frac{(o_j - \mu)}{|| (o_j - \mu) ||})\; |\; o_i \in \mathcal{O} \},
\end{equation}
where $o_i$ is the $i$-th camera center of database images, and $\mu$ is the location the primitive $p$.

\textbf{(3) Geometry Consistency}:
The primitives with smaller multi-view geometric errors and aligned with the scene surface are more reliable for localization.
To measure this term, we use the multi-view 3D distance error as the criteria, defined as:
\begin{equation}
    \text{Geo}(p) = \min(2,\; \frac{{tr}}{\text{Mean}(\{dist_i\})}) + \min(2,\; \frac{{tr}}{\text{Std}(\{dist_i\})}),
\end{equation}
where $\textit{tr}$ is the distance error threshold parameter and $dist_i=\mu - d_i$ is the distance between the primitive position and its corresponding surface point observed by $i$-th camera.

Based on the above-mentioned criteria terms, we compute the final saliency score $\mathcal{H}(p)$ for all key Gaussian primitives:
\begin{equation}
    \mathcal{H}(p) = 2 \cdot \text{Sig}(p) + \min(2,\; \text{Gen}(p)) + \text{Geo}(p),
    \label{eq:landmark_score}
\end{equation}
where the constant coefficients and $\min(\cdot)$ are used for balance the weights for different terms.

\myvspace\noindent\textbf{3D Landmarks Selection.}
To select $N$ effective and non-redundant 3D landmarks for visual localization, we apply a saliency-based greedy search algorithm, as illustrated in~\cref{alg:select_gaussian}.
Our goal is to maximize the overall saliency score of selected landmarks and simultaneously ensure those landmarks cover the reconstructed scene evenly.
This purpose is to enable the query cameras at different locations can observe the selected landmarks as much as possible.

Specifically, the landmark selection process starts by selecting the primitive with the highest saliency score as the initial landmark in the selected set $\mathcal{S}$. 
Next, we search for the primitive with the highest saliency score that is located at a distance greater than the search radius from any of the landmarks already in $\mathcal{S}$. If a suitable primitive is found, it is added to $\mathcal{S}$. If no suitable primitive is found, the search radius is reduced until a suitable primitive is identified. The process continues until the size of the selected set $\mathcal{S}$ reaches the target number $N$, at which point the greedy search stops. 

\subsection{Key Gaussian Primitive Regularization}
\label{subsec:regularization}
In the incremental initialization of Gaussian primitives (\cref{subsec:3dgs}), the primitives initialized by the 2D keypoints are designated as key Gaussian primtives.
However, due to the lack of multi-view geometry optimization, the centers of those key Gaussian primitives are not accurately aligned with the surface and can shift away from their original 3D centers. 
This drift occurs as the primitives adjust to better appearance fitting. 
Such geometry errors can result in inaccurate 3D landmark learning since the misaligned primitives can introduce errors in localization.
To address this, we apply position and scale regularization to the key Gaussian primitives, as illustrated in the middle of~\cref{fig:reconstruction}).
During the differentiable optimization process, we prevent the centers of key Gaussian primitives from being updated to preserve their original alignment. Additionally, we use the 2D keypoint score map generated by~\cite{superpoint} to guide a soft isotropy and scale regularization, which penalizes the scaling parameters as follows:
\begin{equation}
    \mathcal{L}_{reg} = \sum_{\textbf{s},m \in \mathcal{P}_k} | \textbf{s} - \delta \cdot (1 - m) |,
    \label{eq:reg}
\end{equation}
where $\textbf{s}$ and $m$ are the 3D scale vector of the Gaussian primitive and its 2D keypoint score, $\mathcal{P}_k$ is the set of key Gaussian primitive set, and $\delta$ is the scale threshold for key Gaussian primitive.

This regularization term has two purposes.
The first one is to control the 3D scale of the 3D key Gaussian primitives based on its 2D keypoint score reducing their influence during rasterization. The other one is to ensure that key primitives remain isotropic by enforcing a uniform scale constraint across all three axes.

\subsection{Objective Functions}
\label{subsec:loss}
We adopt five different objective functions to optimize our system.

\myvspace\noindent\textbf{Reconstruction Loss.}
Similar to the~\cite{kerbl3Dgaussians,monogs}, for each pixel with ground truth depth and color, we apply the reconstruction losses between the rendered value and ground truth value measured by the camera.
The color and depth reconstruction loss is shown in the following:
\begin{equation}
    \mathcal{L}_{c} = \sum_{i=1} |\hat{I}(i) - I(i)|, \quad \mathcal{L}_{d} = \sum_{i=1} |\hat{D}(i) - D(i)|,
\end{equation}
where $\hat{I}$ and $\hat{D}$ are the rendered color and depth, which are computed by~\cref{eq:render_rgb_depth}.

\myvspace\noindent\textbf{3D Landmark Loss.}
As shown in~\cref{subsec:3dgs}, each Gaussian primitive is associated with a 3D landmark probability score, $a$.
Due to the lack of 3D information, we can not directly optimize the 3D landmark probability score of the primitive.
So, to optimize the probability score, we apply loss between the rendered 2D probability score map $\hat{A}$ (\cref{eq:render_score}) and the 2D probability score map $A$ generated by the~\cite{superpoint}.
The loss is calculated on all rendered pixels:
\begin{equation}
    \mathcal{L}_{m} = \sum_{i=1} \texttt{BCE}(\hat{A}(i), A(i)),
\end{equation}
where $\texttt{BCE}(\cdot)$ is the binary cross-entropy loss.

\myvspace\noindent\textbf{Descriptor Distillation Loss.}
We use the surface-aware descriptor distillation loss term to optimize the scene-specific multi-resolution feature and 3D feature decoder, which is defined as follows:
\begin{equation}
    \mathcal{L}_{dis} = \sum_{x\in \mathcal{X}} |1 - \cos(\mathcal{M}_{f}(\mathcal{E}(x;\Theta)), \mathcal{T}(x;\mathcal{V}))|,
\end{equation}
where $\mathcal{X}$ is the sampled point set on the scene surface manifold, and $\cos(\cdot)$ denotes the cosine similarity of two descriptors.

\myvspace\noindent\textbf{Regularization Loss.}
To mitigate the anisotropic (arrow-shaped) Gaussian primitives and limit the influence of key Gaussian primitives, we adopt the scale and isotropy regularization loss term $\mathcal{L}_{reg}$ (\cref{eq:reg}) introduced in~\cref{subsec:regularization}.

So, the final loss function is presented as follows:
\begin{equation}
\mathcal{L} = \lambda_{1} \cdot \mathcal{L}_{c} + \lambda_{2} \cdot \mathcal{L}_{d} + \lambda_{3} \cdot \mathcal{L}_{m} + \lambda_{4} \cdot \mathcal{L}_{dis} + \lambda_{5} \cdot \mathcal{L}_{reg},
\label{eq:all_loss}
\end{equation}
where \{$\lambda_{i}$\} are the weights for each optimization component.

\subsection{Localization}
\label{subsec:localization}
Once we obtain the reconstructed 3D Gaussian primitives, we can estimate the 6-DoF pose of the query image during the inference stage.
For the query image, $I_q$, we first extract the 2D keypoints with descriptors $\{f_i\}$ via the pre-trained 2D CNN model, SuperPoint~\cite{superpoint}.
For the 3D implicit map, we use all key Gaussian primitives or selected 3D landmarks inside the frustum of the reference image for feature matching.
The reference image in the reconstruction database is obtained via image retrieval with the image-level descriptor~\cite{netvlad}, and the 3D descriptors $\{g_i\}$ of the selected primitives are computed via the scene-specific 3D feature decoder and multi-resolution parametric encoding~\cref{eq:decode_feat}.
To estimate the 2D-3D correspondence, we used the cosine similarity between the 2D and 3D descriptors, $\{f_i\}$ and $\{g_i\}$, as the measure criteria.
So, we can use the RANSAC+PnP~\cite{ransac,epnp} algorithms to estimate the 6-DoF pose based on the estimated correspondences.
\section{Experiments}
\label{sec:exp}
In this section, we first introduce the used datasets and provide the implementation details of our approach.
Then, we evaluate our visual localization and rendering performance.
When compared with multiple baselines, we highlight the best two results with different colors.

\subsection{Datasets}
We evaluate the performance of our method on a variety of scenes from two commonly used datasets.
The Replica~\cite{julian:2019:replica} and 12-Scenes~\cite{12scenes} both contain high-quality RGB-D sequences of various indoor scenes.
For the Replica dataset, we take 8 synthetic scenes provided by~\cite{zhi2021semantic-nerf}.
Each scene contains two sequences, and each sequence includes 900 RGB-D frames.
Following~\cite{uncertainty-loc}, we use the first sequence as the training set, and the second sequence is used for evaluation.
For the 12-Scenes dataset, each scene contains different numbers of RGB-D sequences. We follow the common setting~\cite{uncertainty-loc,12scenes}, the first sequence is used for evaluation and others are used for the training set.

\subsection{Implementation Details}
For the incremental Gaussian initialization process, we project all pixels with 2D keypoints score larger than a threshold $0.005$ into 3D space for key Gaussian primitive initialization.
For the remaining pixels, we random sample $1/64$ points from each keyframe for initialization.
The 3D feature decoder is a 4-layer MLP with 128 hidden units, and the final output dimension of the descriptor is set to 256.
The finest resolution of multi-resolution parametric encoding is 6cm.
The learning rates of position, color, opacity, 3D landmark probability, scale vector, and rotation for each primitive are set to 1.6e-4, 2.5e-3, 0.05, 0.05, 0.001, and 0.001.
Besides, the learning rates of encoding parameters and 3D feature decoder are both set to 1e-3.
We use Adam optimizer for the optimization of Gaussian primitives, multi-resolution encoding parameters, and 3D feature decoder.
For the weight of objective functions in~\cref{eq:all_loss}, we set $\{\lambda_1,\lambda_2,\lambda_3,\lambda_4,\lambda_5\}$ to 1.0, 0.5, 1.0, 1.0, and 0.01, respectively.
After each incremental Gaussian initialization, we random sample 5 frames to construct the keyframes for optimization with 10 iterations.
When we finish the primitive initialization and geometry optimization, we perform additional appearance refinement with 30k iterations.
We adopt the same density strategy in the original 3DGS~\cite{kerbl3Dgaussians}.

\begin{table*}[h]
\centering
\caption{Visual localization performance on Replica dataset. We report median translation and rotation errors (cm, degree).}
\setlength{\tabcolsep}{7.5pt}
\begin{tabularx}{\linewidth}{lcccccccc}
\toprule
Method & Room 0 & Room 1 & Room 2 & Office 0 & Office 1 & Office 2 & Office 3 & Office 4  \\
\midrule
SCRNet~\cite{scrnet} & 2.05 / 0.33 & 1.84 / 0.34 & 1.31 / 0.26 & 1.69 / 0.34 & 2.10 / 0.52 & 2.21 / 0.41 & 2.13 / 0.37 & 2.25 / 0.43 \\
SCRNet-ID~\cite{scrnet-id} & 2.33 / 0.28 & 1.83 / 0.35 & 1.78 / 0.29 & 1.79 / 0.37 & 1.65 / 0.42 & 2.07 / 0.37 & 1.79 / 0.28 & 2.42 / 0.35 \\
SRC~\cite{src} & 2.78 / 0.54 & 1.92 / 0.35 & 2.97 / 0.63 & 1.45 / 0.30 & 2.07 / 0.53 & 2.53 / 0.51 & 3.44 / 0.63 & 4.84 / 0.90\\
NeRF-SCR~\cite{uncertainty-loc} & 1.53 / 0.24 & 1.96 / 0.31 & 1.34 / 0.22 & 1.61 / 0.35 & 1.54 / 0.44 & \cellnd 1.69 / 0.33 & 2.40 / 0.38 & 1.69 / 0.32 \\
PNeRFLoc~\cite{pnerfloc} & \cellnd 1.00 / 0.21 & \cellnd 1.32 / 0.28 & \cellnd 1.43 / 0.29 & \cellnd 0.72 / 0.15 & \cellnd 1.08 / 0.28 & 1.71 / 0.37 & \cellnd 2.39 / 0.30 & \cellnd 1.63 / 0.32 \\
Ours & \cellfs 0.53 / 0.10 & \cellfs 1.06 / 0.20 & \cellfs 1.05 / 0.22 & \cellfs 0.42 / 0.08 & \cellfs 0.85 / 0.21 & \cellfs 1.25 / 0.24 & \cellfs 1.30 / 0.22 & \cellfs 1.10 / 0.19 \\
\bottomrule
\end{tabularx}
\label{tab:replica_pose}
\end{table*}

\begin{table*}[h]
\centering
\setlength{\tabcolsep}{3.4pt}
\caption{Visual localization performance on 12-Scenes dataset. We report median translation and rotation errors (cm, degree).}
\begin{tabularx}{\linewidth}{llccccccccccc}
\toprule
Scenes & \multicolumn{2}{c}{Apartment 1} & \multicolumn{4}{c}{Apartment 2} & \multicolumn{4}{c}{Office 1} & \multicolumn{2}{c}{Office 2} \\
\cmidrule(r){1-1} \cmidrule(r){2-3} \cmidrule(r){4-7} \cmidrule(r){8-11} \cmidrule(r){12-13}
Method & kitchen & living & bed & kitchen & living & luke & gates362 & gates381 & lounge & manolis & 5a & 5b \\
\midrule
SCRNet~\cite{scrnet} & 2.3 / 1.3 & 2.4 / 0.8 & 3.3 / 1.5 & 2.1 / 1.0 & 4.2 / 1.8 & 4.4 / 1.4 & 2.6 / 0.8 & 3.4 / 1.4 & 2.7 / 0.9 & 1.8 / 1.0 & 3.6 / 1.5 & 3.4 / 1.2 \\
SCRNet-ID~\cite{scrnet-id} & 2.6 / 1.4 & 2.0 / 0.8 & 2.0 / 0.8 & 1.8 /  0.9 & 3.0 / 1.2 & 3.7 / 1.3 & 2.1 / 1.0 & 2.9 / 1.2 & 3.4 / 1.1 & 2.6 / 1.2 & 3.3 / 1.2 & 3.8 / 1.3 \\
NeRF-SCR~\cite{uncertainty-loc} & \cellnd 0.9 / 0.5 & 2.1 / 0.6 & \cellnd 1.6 / 0.7 & 1.2 / 0.5 & 2.0 / 0.8 & \cellnd 2.6 / 1.0 & 2.0 / 0.8 & \cellnd 2.7 / 1.2 & \cellnd 1.8 / 0.6 & 1.6 / 0.7 & \cellnd 2.5 / 0.9 & 2.6 / 0.8 \\
PNeRFLoc~\cite{pnerfloc} & 1.0 / 0.6 & \cellnd 1.5 / 0.5 & \cellfs 1.2 / 0.5 & \cellfs 0.8 / 0.4 & \cellnd1.4 / 0.5 & 8.1 / 3.3 & \cellnd1.6 / 0.7 & 8.7 / 3.2 & 2.3 / 0.8 & \cellnd 1.1 / 0.5 & X & \cellnd 2.8 / 0.9 \\
Ours & \cellfs 0.8 / 0.4 &  \cellfs 1.1 / 0.4 &  \cellfs 1.2 / 0.5 &  \cellnd 1.0 / 0.5 &  \cellfs 1.2 / 0.5 &  \cellfs 1.5 / 0.6 &  \cellfs 1.1 / 0.5 &  \cellfs 1.2 / 0.5 &  \cellfs 1.6 / 0.5 &  \cellfs 1.1 / 0.5 &  \cellfs 1.4 / 0.6 &  \cellfs 1.5 / 0.5 \\
\bottomrule
\end{tabularx}
\label{tab:12scenes_pose}
\end{table*}

\begin{table*}[!tp]
\centering
\caption{Novel view synthesis performance on Replica dataset. All scenes are evaluated on the novel views in the test sequence. We outperform PNeRFLoc~\cite{pnerfloc} on the commonly used rendering metrics on all scenes.}
\setlength{\tabcolsep}{7pt}
\begin{tabularx}{\linewidth}{llccccccccc}
\toprule
Method & Metric & Room 0 & Room 1 & Room 2 & Office 0 & Office 1 & Office 2 & Office 3 & Office 4 & Avg. \\
\midrule
\multirow{3}{*}{PNeRFLoc~\cite{pnerfloc}} & \textbf{PSNR} $\uparrow$ &  26.67 & 24.03 & 27.43 & 31.44 & 26.43 & 25.78 & 23.33 & 21.62 & 25.84 \\
                                          & \textbf{SSIM} $\uparrow$ &  0.8730 & 0.8387 & 0.8823 & 0.9355 & 0.8874 & 0.8617 & 0.8026 & 0.7310 & 0.8515 \\
                                          & \textbf{LPIPS} $\downarrow$ &  0.1228 & 0.2231 & 0.2603 & 0.1026 & 0.1397 & 0.1704 & 0.4078 & 0.5259 & 0.2440 \\
\midrule
\multirow{3}{*}{Ours} & \textbf{PSNR} $\uparrow$ & 27.55 & 27.48 & 30.55 & 35.756 & 33.62 & 27.34 & 29.86 & 28.99 & 30.14\\
                      & \textbf{SSIM} $\uparrow$ & 0.8873 & 0.8885 & 0.9396 & 0.9674 & 0.9508 & 0.9158 & 0.9407 & 0.9171 & 0.9259 \\
                      & \textbf{LPIPS} $\downarrow$ & 0.1034 & 0.1106 & 0.0719 & 0.0345 & 0.0624 & 0.0972 & 0.0659 & 0.1024 & 0.0810\\
\bottomrule
\end{tabularx}
\label{tab:replica_rendering}
\end{table*}

\subsection{Evaluation of Localization}
In this part, we show the localization performance on Replica~\cite{julian:2019:replica} and 12-Scenes~\cite{12scenes} datasets. 
Following the common setting~\cite{uncertainty-loc}, we take the images in the second sequence of Replica and the first sequence of 12-Scenes as the test set.
The compared baseline approaches, evaluation metrics, and results of this part are shown in the following.

\myvspace\noindent\textbf{Baseline and Metrics.}
For the evaluation of localization, we follow the setting of~\cite{uncertainty-loc} and take five approaches as our compared baselines, including four regression-based methods: SCRNet~\cite{scrnet}, SCRNet-ID~\cite{scrnet-id}, SRC~\cite{src}, NeRF-SCR~\cite{uncertainty-loc}, and one recent feature-based method PNeRFLoc~\cite{pnerfloc}.
The values of PNeRFLoc~\cite{pnerfloc} are the PnP results obtained with their released codes, and the values of other compared methods are taken from the~\cite{uncertainty-loc}.
To measure the localization accuracy of the different approaches, we use the commonly used relative translation error~\cref{eq:t_error} and relative rotation error~\cref{eq:r_error} as the evaluation metrics:
\begin{equation}
  \Delta t = ||t - \hat{t}||_2,
  \label{eq:t_error}
\end{equation}
\begin{equation}
  \Delta R = \arccos((\text{Tr}(R^T \hat{R}) - 1) / 2),
  \label{eq:r_error}
\end{equation}
where $t$ and $R$ are the ground-truth translation and rotation, respectively, and $\hat{t}$ and $\hat{R}$ are the estimated ones. And $\text{Tr}(\cdot)$ is the trace of the matrix.

\myvspace\noindent\textbf{Qualitative and Quantitative Results.}
To validate the effectiveness of our method, we perform per-frame pose estimation for the query images inside the test set.
The quantitative localization results of Replica~\cite{julian:2019:replica} and 12-Scenes~\cite{12scenes} datasets are shown in~\cref{tab:replica_pose} and \cref{tab:12scenes_pose}, respectively.
As can be seen from~\cref{tab:replica_pose}, compared with the previous best baseline, PNeRFLoc~\cite{pnerfloc}, we achieved the best localization results on all scenes in the Replica dataset with the lowest translation and rotation errors.

For the localization results in the challenging 12-Scenes dataset~\cite{12scenes}, the performance of the point-based neural localization approach, PNeRFLoc~\cite{pnerfloc}, is not as good as in the Replica dataset.
Compared with the NeRF-based scene coordinate regression approach, NeRF-SCR~\cite{uncertainty-loc}, PNeRFLoc shows worse performance on two scenes with translation errors larger than 5cm and rotation errors larger than 3 degrees (scene \texttt{luke} and \texttt{gates381}).
Besides, PNeRFLoc totally failed on scene \texttt{5a}, which is denoted as `X'.
Compared with those baselines, our approach has achieved excellent localization results, achieving the best localization performance in 11 scenes.
Except for the scene \texttt{Apt2/kitchen}, our performance is slightly lower than~\cite{pnerfloc}. 
Among these 12 scenes, our approach can achieve stable localization results.
The reported median errors for all scenes,  the translation, and the rotation of our method are less than 1.6 cm and 0.6 degrees, respectively.
Our approach can achieve more competitive performance localization results among different scenes than the baselines.

\subsection{Evaluation of Rendering}
In this part, we evaluate the performance of novel view rendering on Replica~\cite{julian:2019:replica} dataset. 
Following the common setting~\cite{uncertainty-loc}, we take the images in the second sequence as the test set.
The compared baseline methods, evaluation metrics, and results of this part are shown below.

\myvspace\noindent\textbf{Baseline and Metrics.}
For the evaluation of novel view rendering, we take the recent neural-based visual localization approach, PNeRFLoc~\cite{pnerfloc}, as the compared baseline.
The values of PNeRFLoc~\cite{pnerfloc} are obtained with their released codes.
To evaluate the rendering performance, we adopt the commonly used metrics, Peak Signal-to-Noise Ratio (PSNR), Structural Similarity Index Measure (SSIM), and Learned Perceptual Image Patch Similarity (LPIPS).

\begin{figure*}[h]
  \centering
  \scriptsize
  \setlength{\tabcolsep}{1.5pt}
  \newcommand{\sz}{0.19}
  \begin{tabular}{lccccc}    
    \makecell{\rotatebox{90}{PNeRFLoc~\cite{pnerfloc}}} &
    \makecell{\includegraphics[width=\sz\linewidth]{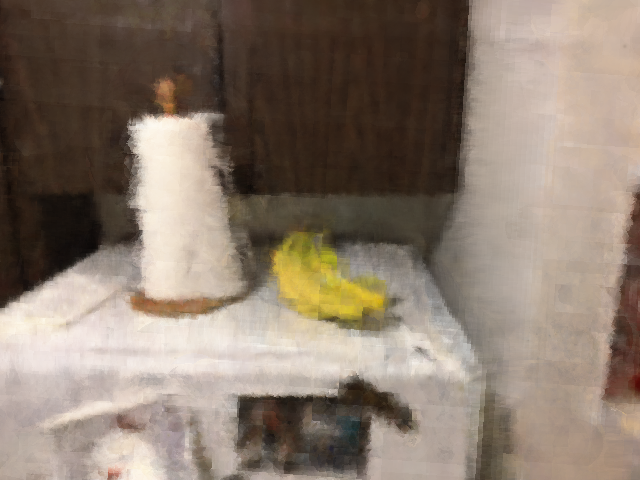}} &   
    \makecell{\includegraphics[width=\sz\linewidth]{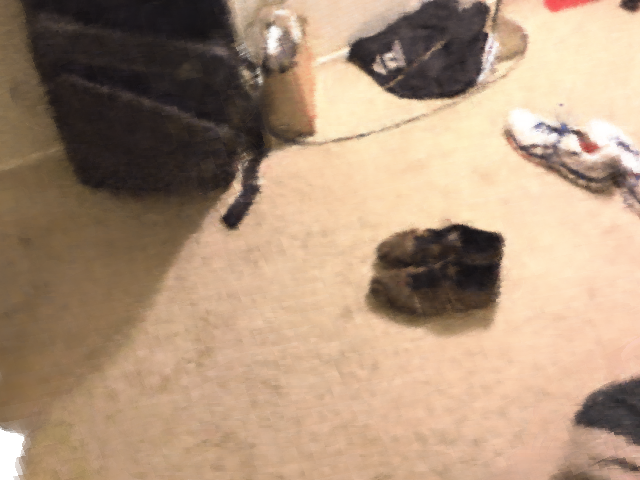}} &   
    \makecell{\includegraphics[width=\sz\linewidth]{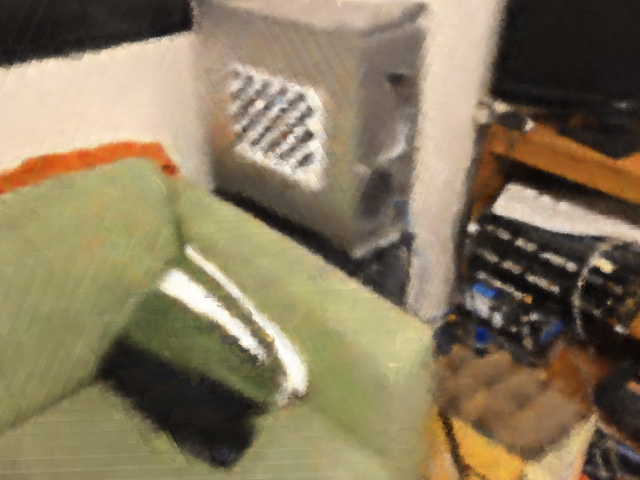}} &   
    \makecell{\includegraphics[width=\sz\linewidth]{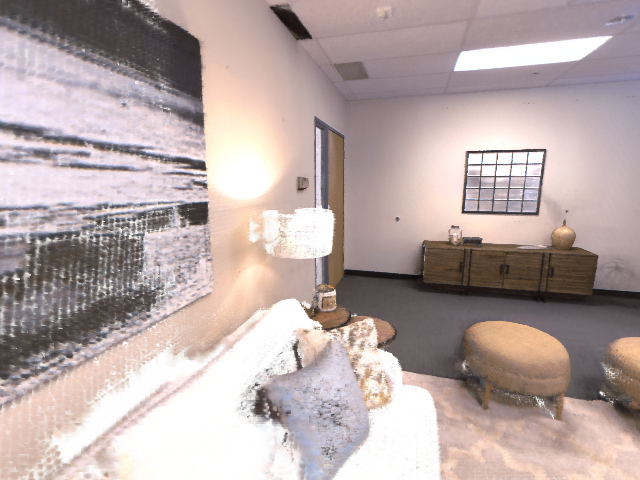}} &   
    \makecell{\includegraphics[width=\sz\linewidth]{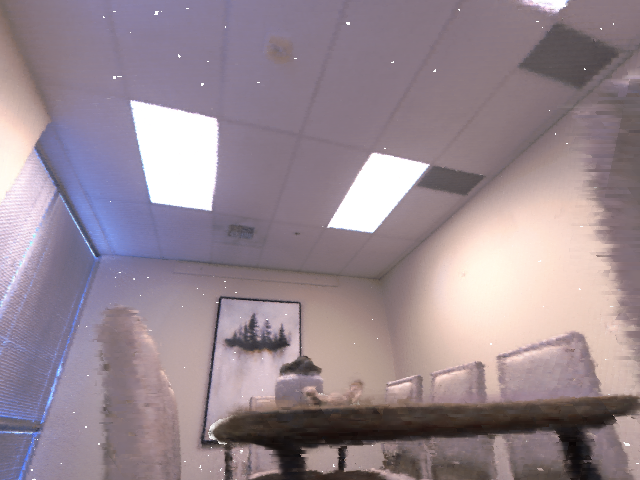}} \\   
    \makecell{\rotatebox{90}{Ours}} &
    \makecell{\includegraphics[width=\sz\linewidth]{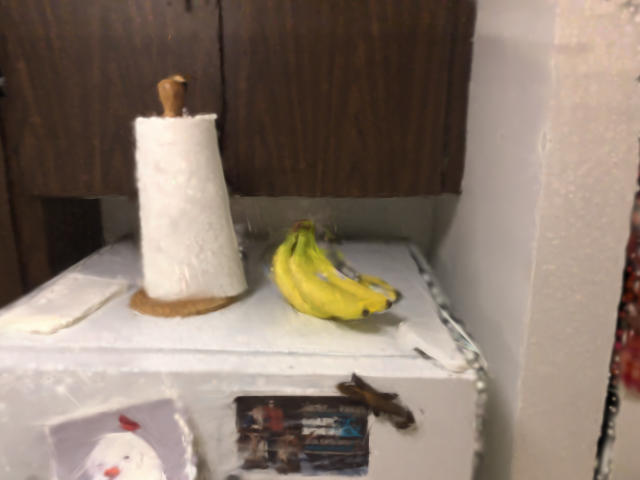}} &   
    \makecell{\includegraphics[width=\sz\linewidth]{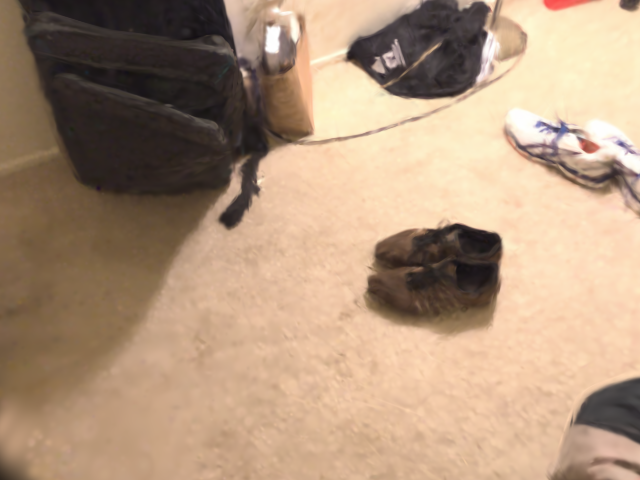}} &   
    \makecell{\includegraphics[width=\sz\linewidth]{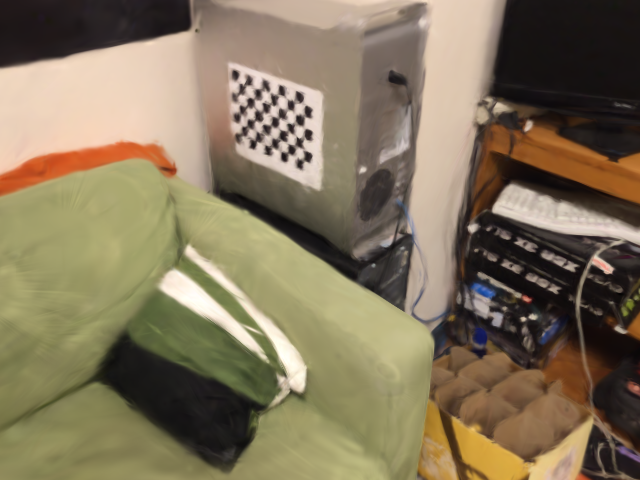}} &   
    \makecell{\includegraphics[width=\sz\linewidth]{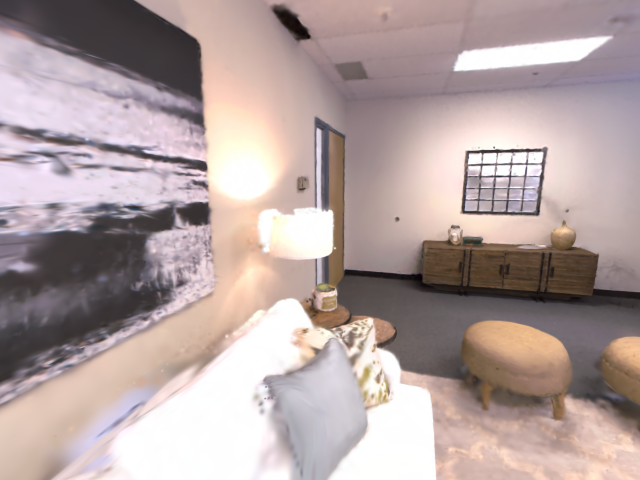}} &   
    \makecell{\includegraphics[width=\sz\linewidth]{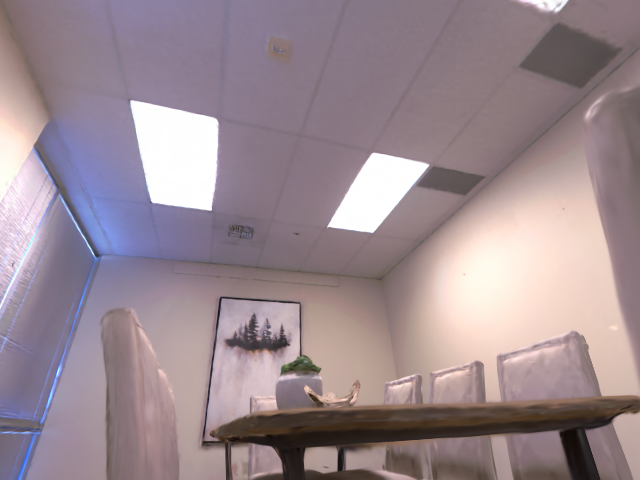}} \\   
    \makecell{\rotatebox{90}{Ground Truth}} &
    \makecell{\includegraphics[width=\sz\linewidth]{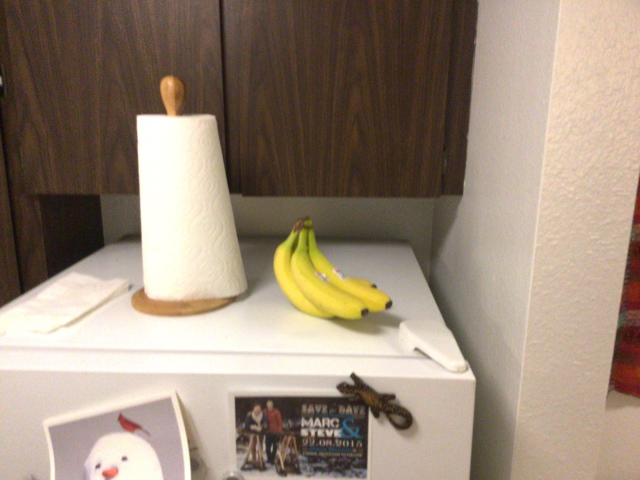}} &   
    \makecell{\includegraphics[width=\sz\linewidth]{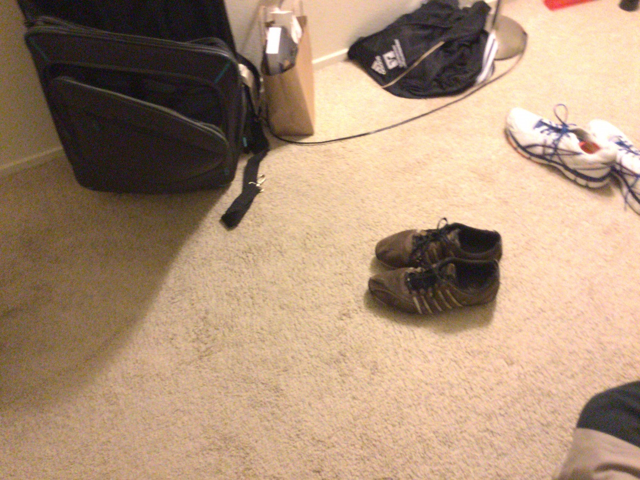}} &   
    \makecell{\includegraphics[width=\sz\linewidth]{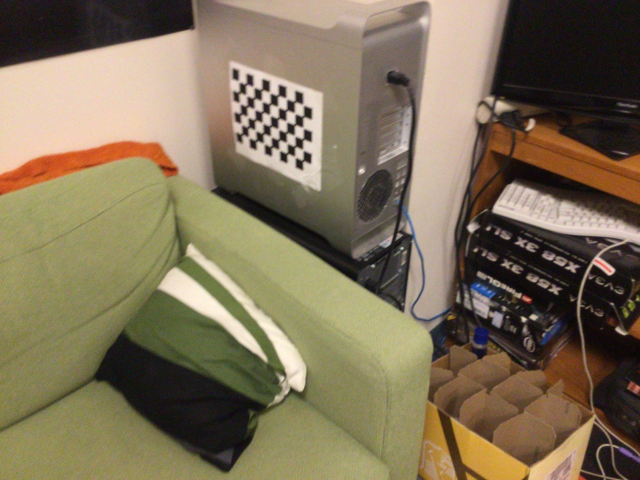}} &   
    \makecell{\includegraphics[width=\sz\linewidth]{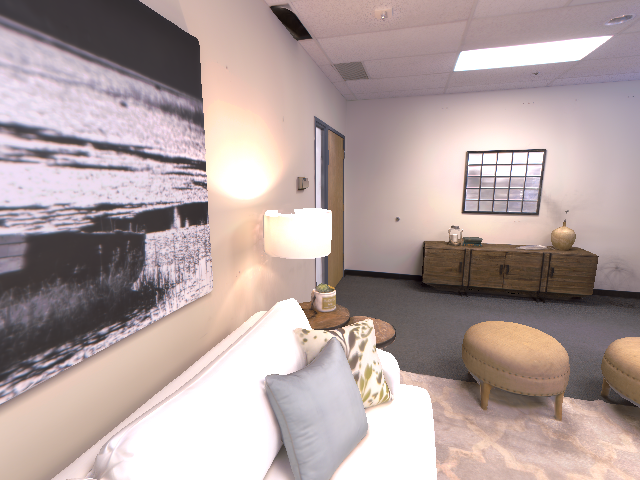}} &   
    \makecell{\includegraphics[width=\sz\linewidth]{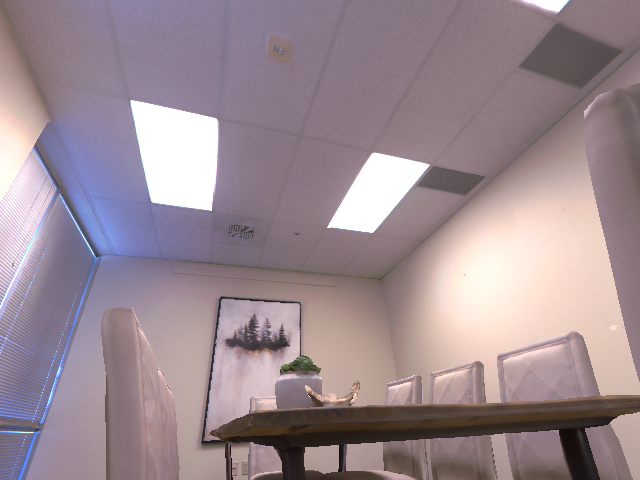}} \\   
    & \tt Apt1/kitchen & \tt Apt2/bed & \tt Of1/gates381 & \tt Room 0 & \tt Room 2 \\
  \end{tabular}
\caption{Visualization of novel view synthesis. We show some novel view rendering results from different scenes. From top to bottom, there are results of PNeRFLoc~\cite{pnerfloc}, ours, and ground truth. Our rendering results are more clear and have less noise information.}
\label{fig:nvs_rendering}
\end{figure*}

\myvspace\noindent\textbf{Qualitative and Quantitative Results.}
We show the novel view rendering performance on each viewpoint in the test set.
The quantitative results of Replica~\cite{julian:2019:replica} dataset are reported in~\cref{tab:replica_rendering}.
The results show that our approach significantly outperforms the compared baseline on all scenes by a substantial margin.
Compared with PNeRFLoc~\cite{pnerfloc}, we achieve the $4.302$, $0.074$, and $0.163$ performance improvement on PSNR, SSIM, and LPIPS metrics, respectively.
Besides, we also show the qualitative novel view rendering results of several selected scenes in~\cref{fig:nvs_rendering}.
From the figure, we can see that our method can generate clearer results than PNeRFLoc~\cite{pnerfloc}.
PNeRFLoc cannot model the high-frequency appearance details of the scene very well and may produce artifacts and blurry rendering results from novel viewpoints.
In contrast, our method can better reconstruct the appearance information of the scene and have better rendering results from novel viewpoints.
As shown in the~\cref{tab:replica_rendering}, higher LPIPS indicates higher perceptual similarity of images rendered by our method.


\begin{table}[!ht]
\centering
\caption{Training time, memory usage, and rendering FPS of different methods on scene \texttt{manolis} from 12-Scenes Dataset.}
\setlength{\tabcolsep}{8pt}
\begin{tabularx}{\linewidth}{lccc}
\toprule
Method & Training Time$\downarrow$ & Memory$\downarrow$ & FPS$\uparrow$\\
\midrule
SCRNet~\cite{scrnet} &  2 days & 165 MB & - \\
NeRF-SCR~\cite{uncertainty-loc}  & 16 hours & - & - \\
PNeRFLoc~\cite{pnerfloc} &  1 hour & 788 MB & 0.23 \\
Ours  &  25 mins & 112 MB & 498 \\
\bottomrule
\end{tabularx}
\label{tab:time_memory}
\end{table}

\subsection{Memory Usage and Time Analysis}
In this part, we measure the memory usage, method training time, and rendering speed of the neural-based localization approach.
The results are shown in~\cref{tab:time_memory}, evaluated in the scene \texttt{manolis}.
Due to SCRNet~\cite{scrnet} can not perform novel view rendering, we do not report it rendering FPS performance.
And NeRF-SCR~\cite{uncertainty-loc} don't release their codes and do not report their memory usage, FPS in their paper. So, we also don't report them in the table.
As can be seen from the table, our method has a faster training speed, smaller model storage, and faster rendering speed.
Scene coordinate regression-based approaches, \textit{e.g.}, SCRNet~\cite{scrnet} and NeRF-SCR~\cite{uncertainty-loc}, need more than ten hours or even days for the convergence of the regression network.
Due to the fast differentiable rasterization process, our approach has a faster convergent speed than the point-based neural implicit approach, PNeRFLoc~\cite{pnerfloc} (25 mins V.S. 1 hour).
Due to the fact that we learn the descriptor from the 3D feature decoder not the storage descriptor for each primitive, our storage memory is 7x less than~\cite{pnerfloc} (112 MB V.S. 788 MB). 
Besides, for the rendering speed (FPS) at the resolution of $640\times480$, our rendering speed is 2k times faster than~\cite{pnerfloc} (498 FPS V.S 0.23 FPS). 
Our approach can not only estimate the accurate 6-DoF pose for the query camera but also perform real-time high-quality novel view rendering performance, which is more suitable for virtual reality and augmented reality.

\subsection{Ablation Studies}
In this part, we conduct ablation studies to validate the effect of each part or design in our system.



\begin{table}[h]
\centering
\caption{Visual localization performance with different descriptor learning settings. We report the median translation and rotation errors (cm, degree) on two selected scenes, \texttt{Room 0} from Replica and \texttt{apt1/kitchen} from 12-Scenes.}
\setlength{\tabcolsep}{6pt}
\begin{tabularx}{\linewidth}{lccc}
\toprule
Case & Description & Room 0 & Apt1/Kitchen\\
\midrule
\texttt{\#}1 & Blending w/ Auto E.D. &  2.97 / 1.69   & 3.90 / 2.69  \\
\texttt{\#}2 & Ours 3D Dec. w/o P.E. &  82.97 / 16.68 & 13.27 / 8.13  \\
\texttt{\#}3 & Ours 3D Dec. w/ P.E. &  6.87 / 4.60  & 7.53 / 3.71  \\
\texttt{\#}4 &  Ours w/o S.D.D &   1.14 / 0.58  &  1.86 / 1.02  \\
\texttt{\#}5 &  Ours Full & \textbf{0.53} / \textbf{0.10} &  \textbf{0.77}  / \textbf{0.43}  \\
\bottomrule
\end{tabularx}
\label{tab:diff_decoder}
\end{table}

\myvspace\noindent\textbf{Effects of Different Descriptor Learning Settings.}
In~\cref{tab:diff_decoder}, we show the performance of using different kinds of approaches to learning descriptors for Gaussian primitive on two selected scenes.
\texttt{\#}1 represents the way used in~\cite{qin2024langsplat}, which uses an auto encoder-decoder (Auto E.D.) to compress the high dimensional feature and perform biased alpha-blending.
\texttt{\#}2 and \texttt{\#}3 represent that using our 3D feature decoder without/with positional encoding, respectively.
\texttt{\#}4 represents our approach with our muti-resolution parametric encoding but without surface-aware descriptor distillation (S.D.D).
We can draw some conclusions from the results in the table.
Comparing \texttt{\#}2, \texttt{\#}3, and \texttt{\#}4, the descriptor obtained by just optimizing a 3D feature decoder is not accurate enough. In large indoor scenes, the representation ability of an MLP is not enough to learn scene features with sufficiently high discrimination ability, and additional parametric encoding needs to be used to assist the descriptor learning of 3D scenes.
Comparing \texttt{\#}1 and \texttt{\#}4, our feature learning method is more accurate than the alpha-blending method.
Comparing \texttt{\#}4 and \texttt{\#}5, we show that paying more attention to the 3D points on the scene geometry surface can lead to better and faster convergence for primitive descriptor learning.
Compared with the baselines, our unbiased 3D descriptor learning approach can achieve better localization results.

\begin{figure}[h]
\centering
\includegraphics[width=0.48\linewidth]{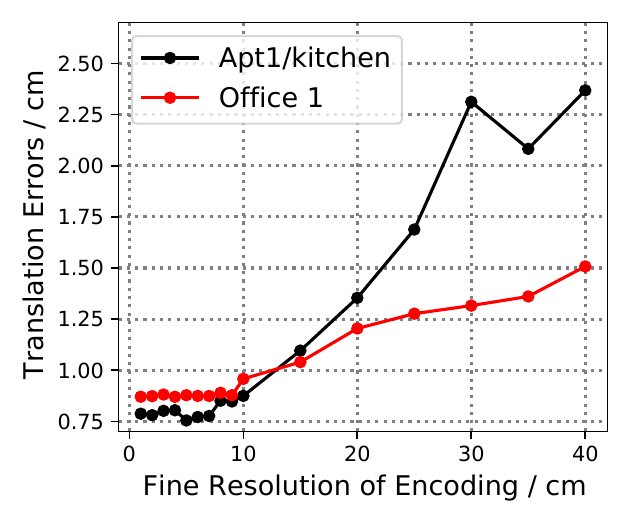}
\includegraphics[width=0.48\linewidth]{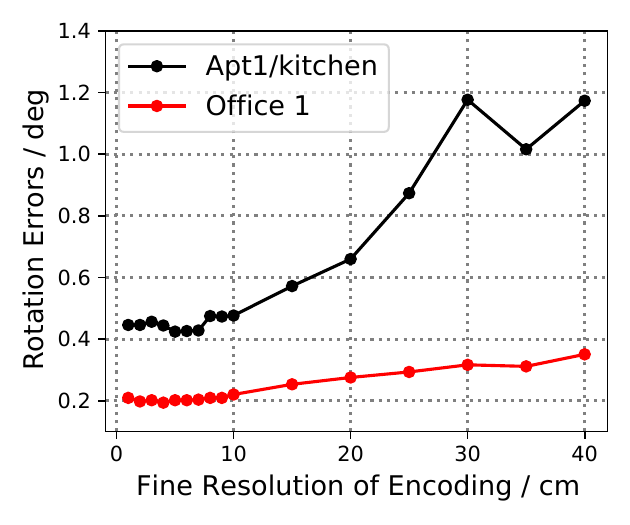}
\caption{Visual localization performance of using different resolutions of parametric encodings. We report median translation and rotation errors (cm, degree) on two selected scenes.}
\label{fig:ablation_diff_hash_resolution}
\end{figure}

\myvspace\noindent\textbf{Different Resolution of Parametric Encoding.}
In~\cref{subsec:distill_feat}, we adopt multi-resolution parameter encoding to enhance the representation ability of the single MLP decoder.
We investigate the performance of the spatial resolution used for encoding feature grids.
The results on two selected scenes (\texttt{Apt1/kitchen} from 12-Scenes and \texttt{Office 1} from Replica) are shown in~\cref{fig:ablation_diff_hash_resolution}.
As shown in the figure, when the resolution of parametric encoding is greater than 20 cm, the localization accuracy will quickly decrease as the resolution increases. When the resolution is below 10 cm, the localization accuracy will converge to a stable value.
So, we set the resolution as 6 cm for all scenes in our experiments.

\begin{figure}[h]
\centering
\includegraphics[width=0.48\linewidth]{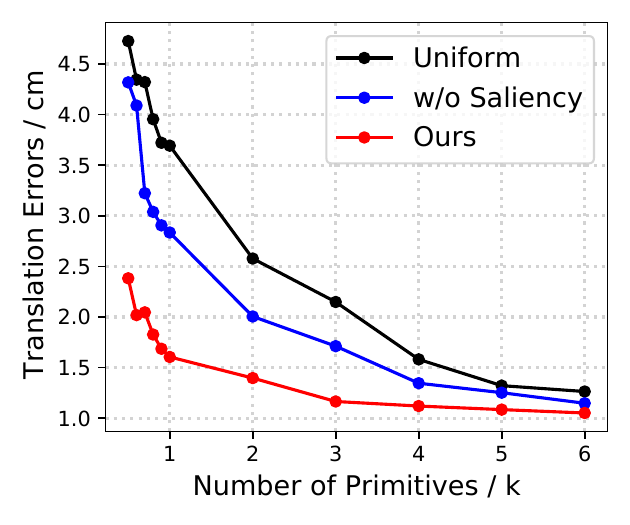}
\includegraphics[width=0.48\linewidth]{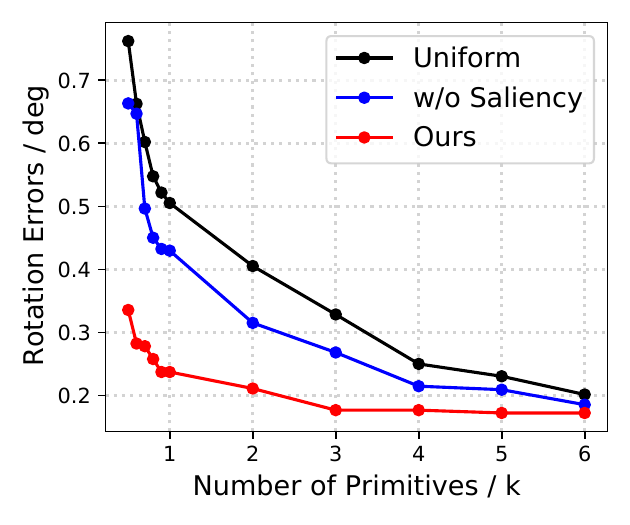}
\label{fig:diff_downsample}
\caption{Visual localization performance of using different numbers of 3D landmark. We report median translation and rotation errors (cm, degree) on the scene of \texttt{Room 0} from the Replica Dataset.}
\end{figure}

\myvspace\noindent\textbf{Effects of Different Numbers of 3D Landmarks.}
We show the localization performance of using the different numbers of selected landmarks.
To validate the effectiveness of our proposed selection algorithm (\cref{subsec:score_select}), we take several comparison baselines into consideration.
The localization results are shown in~\cref{fig:diff_downsample}.
`Uniform' denotes the performance uniform downsample of the Gaussian primitives.
`w/o saliency' denotes the performance achieved by the greedy search method but without our saliency score for optimal selection.
As can be seen from the figure, our proposed landmark selection algorithm can lead to better performance when only a small number of 3D landmarks are used for localization ($\sim$ 500 landmarks).
Those results without using greedy search and saliency score perform worse results for the situation of small primitives.
Compared with the baselines, our 3D landmark selection algorithm can achieve better localization results.

\begin{figure}[h] 
\centering
\includegraphics[width=0.49\linewidth]{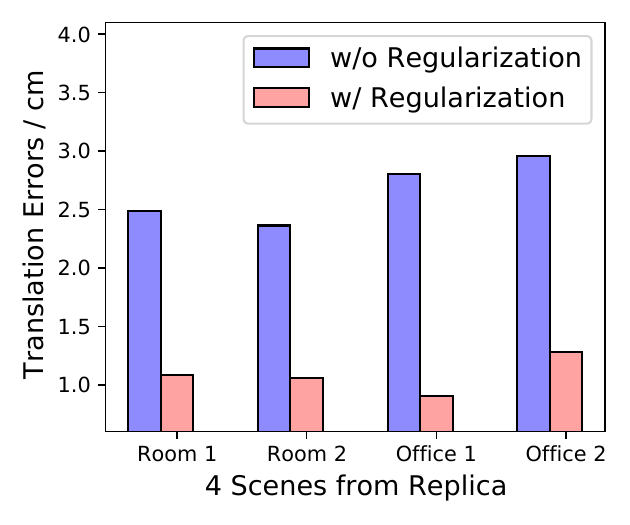}
\includegraphics[width=0.49\linewidth]{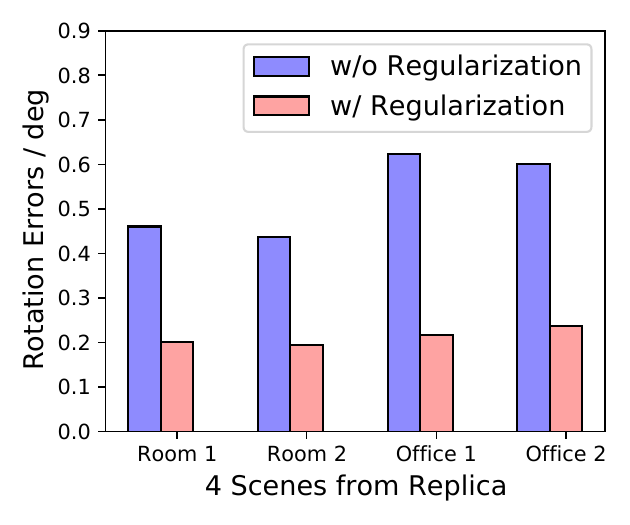}
\caption{Visual localization performance of using primitive regularization terms. We report median translation and rotation errors (cm, degree) on four selected scenes from Replica.}
\label{fig:ablation_regularization}
\end{figure}

\myvspace\noindent\textbf{Effectiveness of Key Gaussian Primitive Regularization.}
We validate the performance of key Gaussia primitive regularization (\cref{subsec:regularization}) used for our visual localization approach on four selected scenes in the Replica dataset~\cite{julian:2019:replica}.
The localization results are shown in~\cref{fig:ablation_regularization}.
`w/o Regularization' and `w/ Regularization' represent the performance of not using and using position and scale regularization terms for key Gaussian primitives, respectively.
Using regularization terms can lead to more consistent and accurate 3D landmark learning.
As can be seen from the figure, with regularization terms, we can achieve better and more stable localization performance improvement.
Across the selected four scenes, using regularization terms can reduce the error by an average of 1.57 cm and 0.32 degrees.

\subsection{AR Applications}

\begin{figure}[h]
  \centering
  \setlength{\tabcolsep}{1.5pt}
  \newcommand{\sz}{0.31}
  \begin{tabular}{lccc}
    &  \tt View 1 &  \tt View 2 & \tt View 3 \\
    \makecell{\rotatebox{90}{\scriptsize{Orginal Images}}} &
    \makecell{\includegraphics[width=\sz\linewidth]{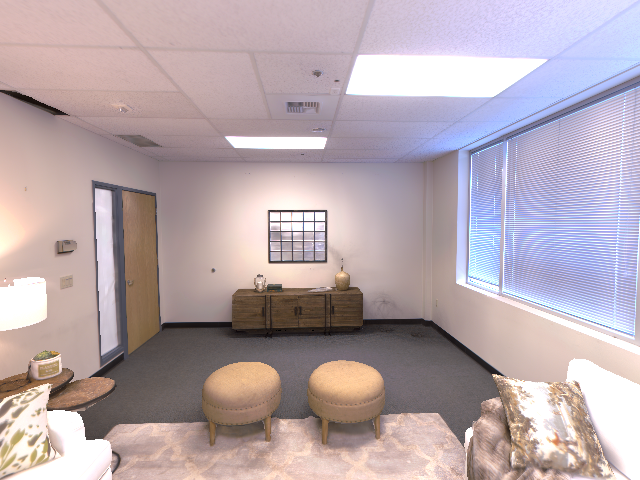}} &
    \makecell{\includegraphics[width=\sz\linewidth]{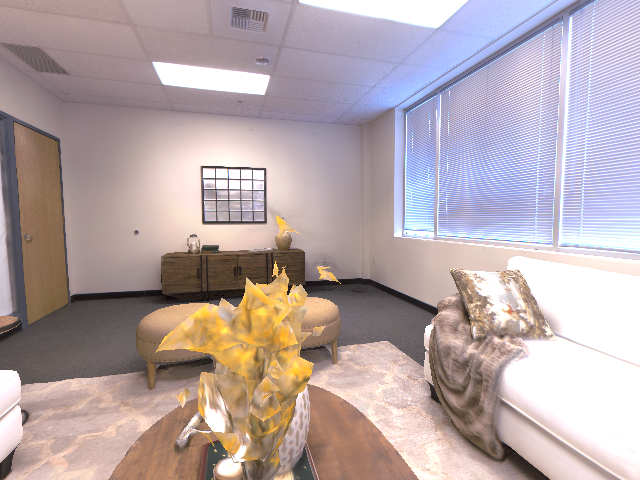}} &
    \makecell{\includegraphics[width=\sz\linewidth]{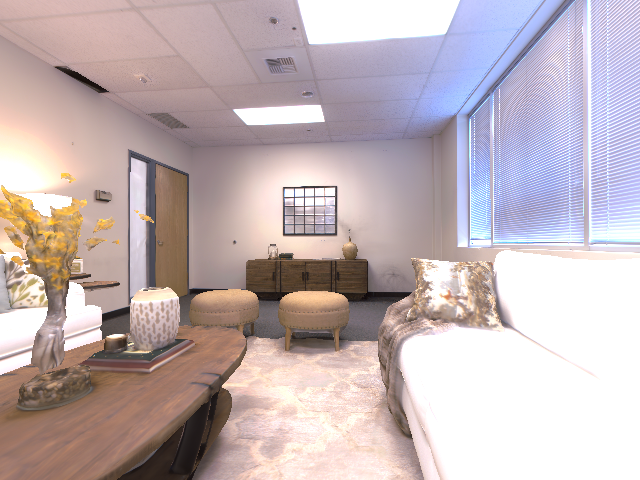}} \\
    \makecell{\rotatebox{90}{\scriptsize{Insert Objects}}} &
    \makecell{\includegraphics[width=\sz\linewidth]{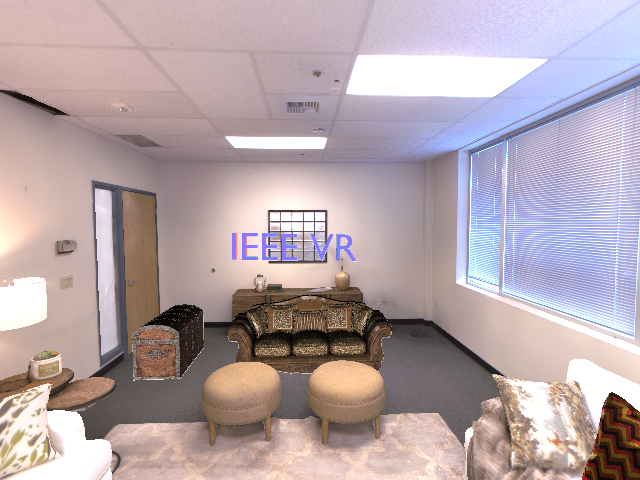}} &
    \makecell{\includegraphics[width=\sz\linewidth]{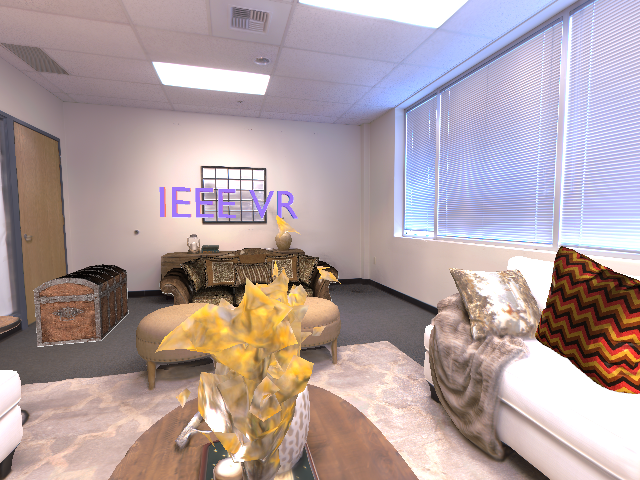}} &
    \makecell{\includegraphics[width=\sz\linewidth]{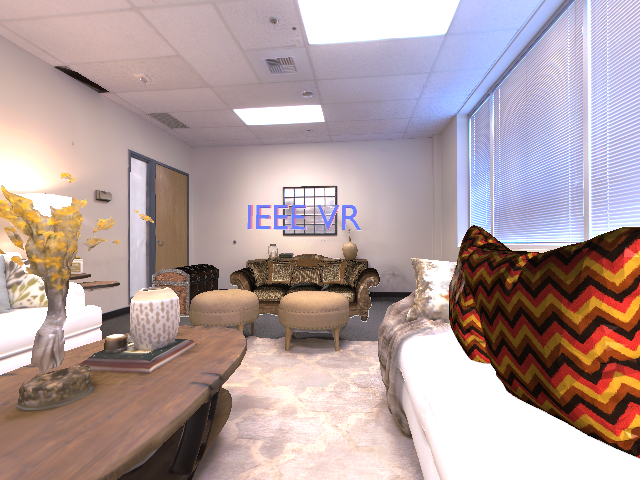}} \\
    \makecell{\rotatebox{90}{\scriptsize{Physical Collision}}} &    
    \makecell{\includegraphics[width=\sz\linewidth]{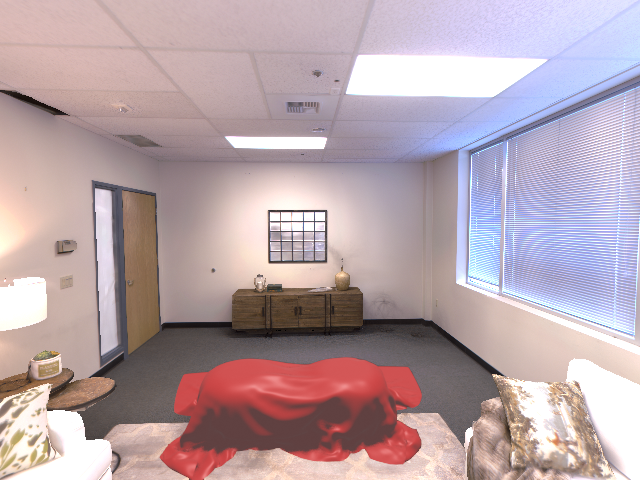}} &
    \makecell{\includegraphics[width=\sz\linewidth]{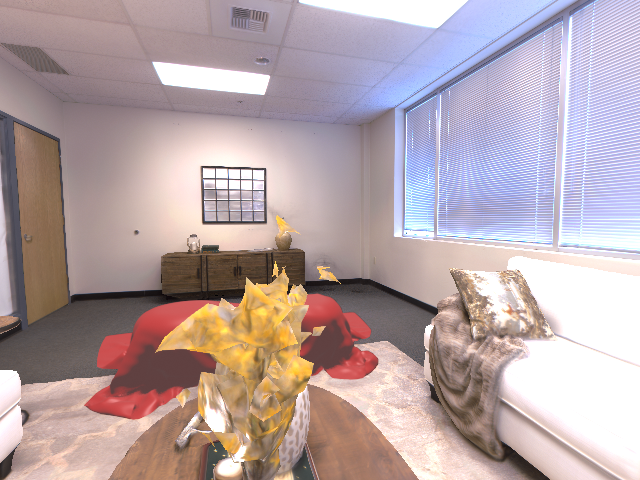}} &
    \makecell{\includegraphics[width=\sz\linewidth]{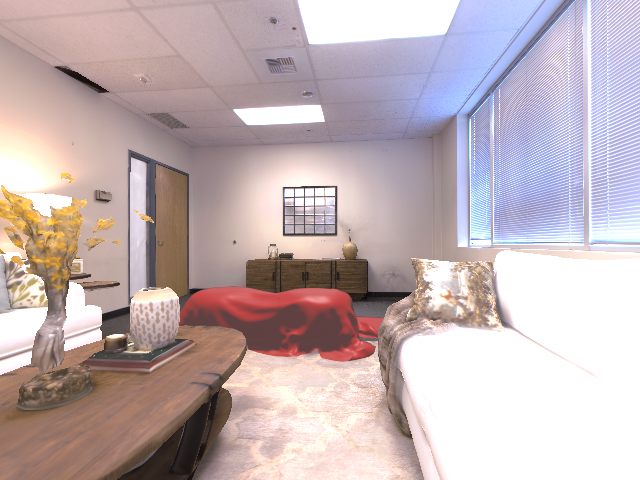}} \\
  \end{tabular}
\caption{The AR demo on the scene \texttt{Room 0} of Replica~\cite{julian:2019:replica}. We show two kinds of augmented reality applications. The first row shows the original RGB image. The second row shows the images with our inserted AR objects and \texttt{IEEE VR} text logo. The third row shows the physical collision between our reconstructed geometry and the virtual blanket.}
\label{fig:ar_demo}
\end{figure}

In this part, we show the applications of our visual localization approach in the AR field.
Our approach can not only estimate the 6-DoF pose of the camera in the reconstructed scene but also perform high-quality novel view rendering based on the Gaussian primitives.
Therefore, our method can be very suitable for AR applications.
We show two different AR applications in~\cref{fig:ar_demo} on scene \texttt{Room 0} from the Replica dataset.
\textbf{(1) Insert Objects}: we virtual AR objects and \texttt{IEEE VR} text logo into real scene.
\textbf{(2) Physical collision}: we place a virtual blanket in the scene and let it fall naturally to simulate physical collision between our reconstructed scene geometry and the virtual blanket.
As can be seen, our approach can perform high-quality rendering and handle the collision and occlusion between real and virtual content very well.
For more AR demo videos, please refer to our supplementary video.
\section{Conclusion and Limitation}
\label{sec:conclusion}
In this paper, we propose \ours, an efficient and novel visual localization approach based on the 3D Gaussian primitives, which is more suitable for AR/VR than traditional localization methods.
Specifically, to compress the scene model for localization, we learn an unbiased 3D descriptor field for reconstructed Gaussian primitives, which is more accurate than previous alpha-blending approaches.
Then, we propose a salient 3D landmark selection algorithm to select more informative primitives for visual localization based on the saliency score of Gaussian primitives, which can reduce the memory and runtime requirements for mobile devices.
Besides, we propose an effective regularization term for key Gaussian primitives to avoid anisotropic shapes and reduce geometric errors, which can lead to stable localization performance improvement.
Extensive experiments on two commonly used datasets have shown the effectiveness and application of our proposed system.

Currently, our proposed approach has two limitations.
The first one is that we need depth information or sparse point clouds to reconstruct the scene.
Our approach is based on 3DGS~\cite{kerbl3Dgaussians}, which needs point cloud to initialize the position of each Gaussian primitive.
The second one is that our method cannot be used for large outdoor scenes, which will increase the number of parameters.
In the future, we will try to use the visual foundation model (\textit{e.g.}, DepthAnything~\cite{depthanything}) to estimate the depth of the RGB image, which can be viewed as prior to replacing the depth sensor and guide the scene reconstruction process.
Besides, considering using a hierarchical representation approach~\cite{hierarchicalgaussians24} to extend our localization approach for large outdoor scenes.



\bibliographystyle{abbrv-doi} 

\bibliography{draft}
\end{document}